%% file: main.tex
\documentclass[10pt,twocolumn,letterpaper]{article}

\usepackage[pagenumbers]{iccv} 


\definecolor{iccvblue}{rgb}{0.21,0.49,0.74}
\usepackage[pagebackref,breaklinks,colorlinks]{hyperref}  
\usepackage{multirow} 

\def\paperID{1136} 
\def\confName{ICCV}
\def\confYear{2025}

\title{Revisiting Monocular 3D Object Detection with Depth Thickness Field}

\author{Qiude Zhang, Chunyu Lin\thanks{Corresponding author.}, Zhijie Shen, Nie Lang, and Yao Zhao\\
Institute of Information Science, Beijing Jiaotong University\\
Beijing Key Laboratory of Advanced Information Science and Network Technology\\
{\tt\small \{zhangqiude, cylin\}@bjtu.edu.cn}}

\begin{document}
\maketitle
\input{sec/0_abstract}    
\input{sec/1_intro}
\input{sec/2_relwork}

\input{sec/3_method}

\input{sec/4_exps}
\input{sec/5_conclusion}
{
    \small
    \bibliographystyle{ieeenat_fullname}
    \bibliography{main}
}
\end{document}

%% file: sec/0_abstract.tex
\begin{abstract}
Monocular 3D object detection is challenging due to the lack of accurate depth. However, existing depth-assisted solutions still exhibit inferior performance, whose reason is universally acknowledged as the unsatisfactory accuracy of monocular depth estimation models. In this paper, we revisit monocular 3D object detection from the depth perspective and formulate an additional issue as the limited 3D structure-aware capability of existing depth representations (\textit{e.g.}, depth one-hot encoding or depth distribution).
To address this issue, we introduce a novel \textbf{Depth Thickness Field} approach to embed clear 3D structures of the scenes. Specifically, we present \textbf{MonoDTF}, a scene-to-instance depth-adapted network for monocular 3D object detection. 
The framework mainly comprises a Scene-Level Depth Retargeting (SDR) module and an Instance-Level Spatial Refinement (ISR) module.
The former retargets traditional depth representations to the proposed depth thickness field, incorporating the scene-level perception of 3D structures. 
The latter refines the voxel space with the guidance of instances, enhancing the 3D instance-aware capability of the depth thickness field and thus improving detection accuracy.
Extensive experiments on the KITTI and Waymo datasets demonstrate our superiority to existing state-of-the-art (SoTA) methods and the universality when equipped with different depth estimation models.
The code will be available.
\end{abstract}

%% file: sec/1_intro.tex
\section{Introduction}
\label{sec:intro}

\begin{figure}[t]
  \centering
  \begin{subfigure}{0.99\linewidth}
    \includegraphics[width=1.0\linewidth]{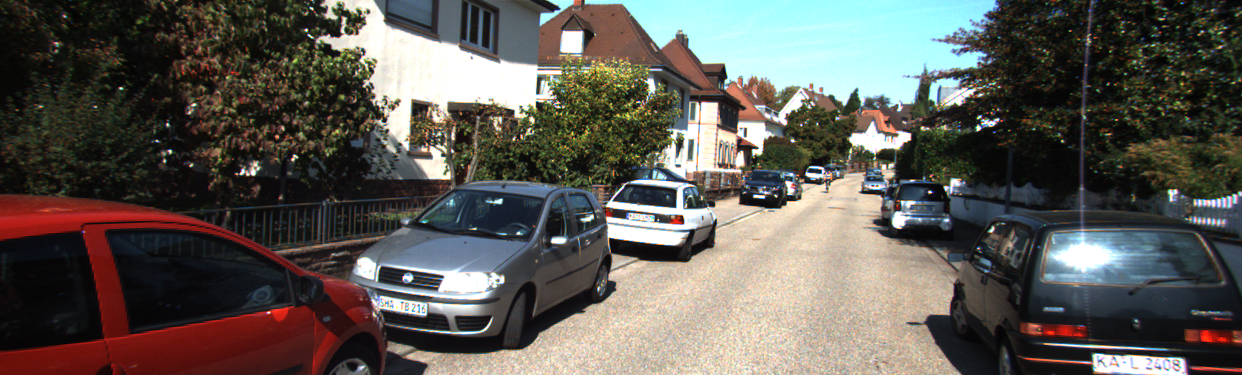}
    \caption*{Input RGB Image}
    \label{fig:a}
  \end{subfigure}
  \begin{subfigure}{0.49\linewidth}
    \flushleft
    \includegraphics[width=0.96\linewidth]{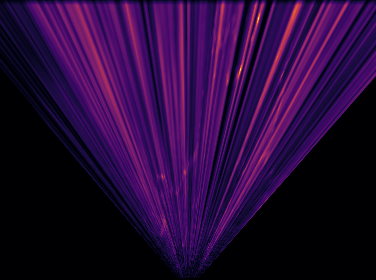}
    \caption{Depth One-Hot Encoding}
    \label{fig:b}
  \end{subfigure}
  \begin{subfigure}{0.49\linewidth}
    \flushright
    \includegraphics[width=0.96\linewidth]{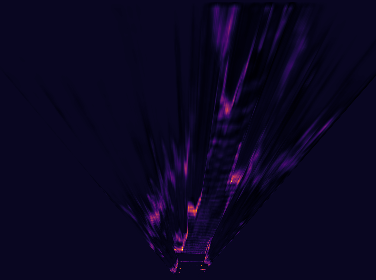}
    \caption{Depth Distribution}
    \label{fig:c}
  \end{subfigure}
  \begin{subfigure}{0.49\linewidth}
    \flushleft
    \includegraphics[width=0.96\linewidth]{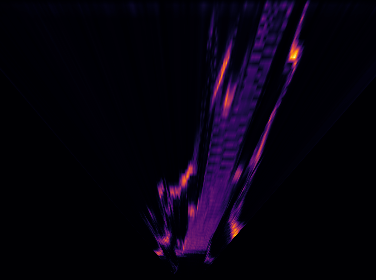}
    \caption{Depth Thickness Field (Ours)}
    \label{fig:d}
  \end{subfigure}
  \begin{subfigure}{0.49\linewidth}
    \flushright
    \includegraphics[width=0.96\linewidth]{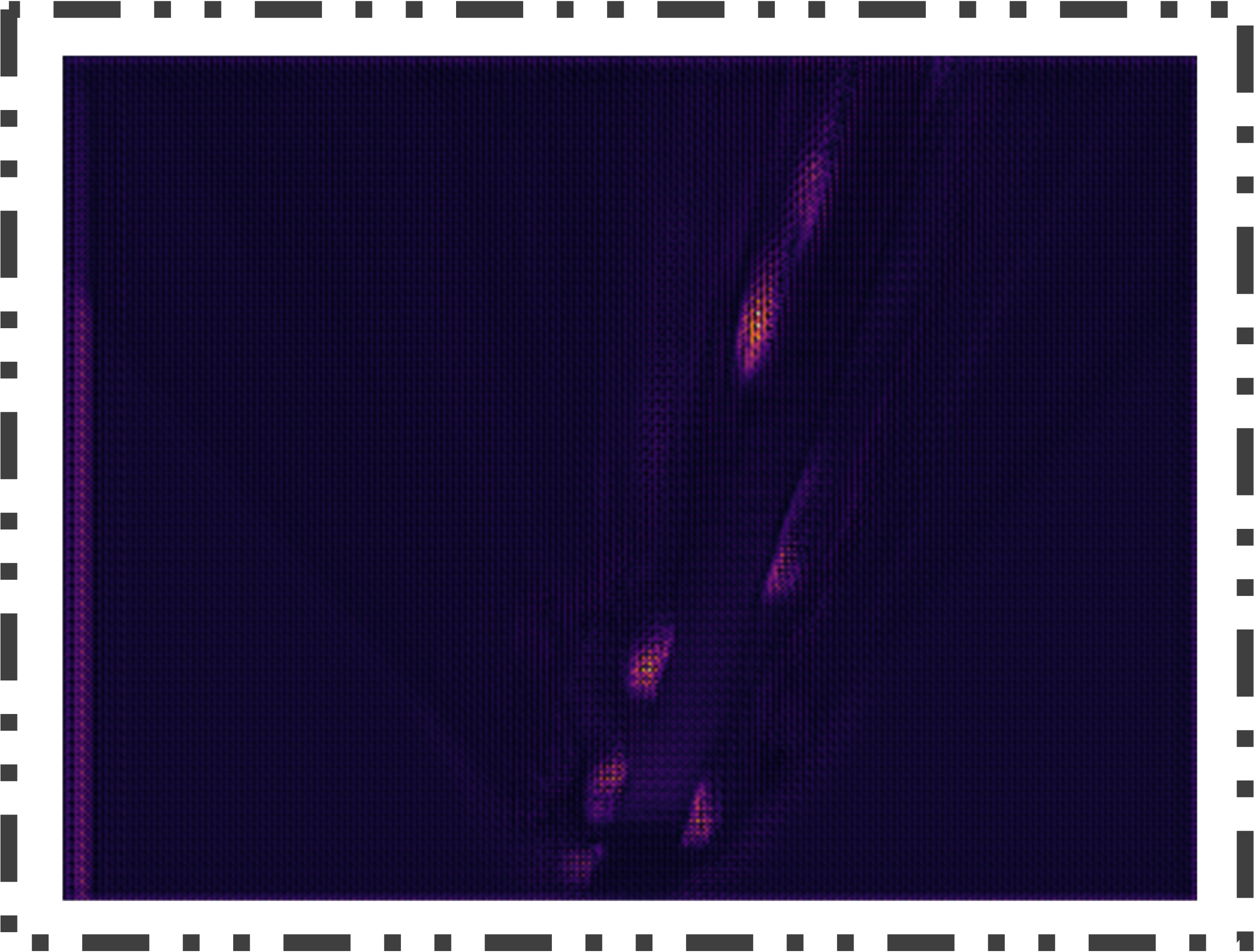}
    \caption*{Instance-Aware Feature (Ours)}
    \label{fig:e}
  \end{subfigure}
   \caption{(a), (b), and (c) show VOXEL features from three different depth representations. (c) provides a more comprehensive and clearer scene structure feature. The last image surrounded by a dashed box represents instance-aware BEV features derived from (c), focusing on instance structure.}
   \label{fig:one}
\end{figure}

3D object detection is critical in computer vision and has a wide range of applications, such as autonomous driving, robotic perception, etc. 
Although the methods based on LiDAR \cite{yan2018second,lang2019pointpillars,shi2020pv} and multi-view images \cite{philion2020lift,huang2021bevdet,li2022bevformer,li2023bevdepth} have achieved promising performance, the heavy dependence on the expensive equipment \cite{chong2022monodistill, ma20233d} restricts the application of these methods. Recently, there have been considerable advancements in monocular methods \cite{peng2024learning,liu2024monotakd,liao2024monodetrnext, peng2022did, xu2023mononerd} and garnered substantial attention from both academic and industrial communities, which give credit to their low cost and ease of deployment.

Monocular 3D object detection can be directly assisted by monocular depth estimation. However, inferring depth from a monocular image is inherently an ill-posed problem that affects the performance of monocular 3D object detection. Early depth-assisted methods \cite{ding2020learning,wang2019pseudo} resort to pre-training an auxiliary depth estimator, \textit{e.g.}, MonoDepth2 \cite{godard2019digging}, to predict depth maps, but cannot achieve satisfactory detection performance. 
Most solutions (\textit{e.g.}, DD3D \cite{park2021pseudo}) attribute it to the limited accuracy of monocular depth estimation models while overlooking the significance of depth representations. 
Recently, benefiting from the large-scale dataset and self-supervised training strategy, monocular depth estimation using large models (\textit{e.g.}, Depth Anything V2 (DAM v2) \cite{yang2024depth} and Metric3D v2 \cite{yin2023metric3d}) exhibit impressive accuracy and generalization. 
An intuitive idea is to leverage them to improve monocular 3D object detection. Unfortunately, as reported in Sec. \textcolor{red}{\ref{ablation}}, the results remain unsatisfactory, which drives us to rethink the well-recognized issue (\textit{i.e.}, limited depth accuracy problem) and revisit this type of approaches. 
In this work, we hold a different idea and contend that this issue also stems from the limited 3D structure awareness of existing depth representations (\textit{e.g.}, depth one-hot encoding or depth distribution).

To address this issue, we propose MonoDTF, a novel scene-to-instance depth-adapted monocular 3D detection network. Different from existing approaches \cite{reading2021categorical, peng2024learning} that adapt Lift-Splat-Shoot (LSS) \cite{philion2020lift} framework and bird’s-eye-view (BEV) space, MonoDTF retargets traditional depth representations in a new formulation: \textbf{Depth Thickness Field (DTF)} to adapt to 3D object detection. The DTF can reduce the impact of erroneous depth estimation and integrate the thickness property for each image pixel in the 3D space. It implicitly describes the thickness along camera rays, producing sharper and more accurate features, as illustrated in Fig. \ref{fig:one}. 
MonoDTF consists of two main components. First of all, a Scene-Level Depth Retargeting (SDR) module is presented to retarget traditional depth representations to the proposed depth thickness field. Compared with previous representations, the DTF embeds clear 3D structure awareness, benefiting in distinguishing different instances and backgrounds.
Then, we sample image features and depth thickness field to establish the voxel space with predefined grids. 
Second, an Instance-Level Spatial Refinement (ISR) module is proposed to refine the voxel space by raising the oversight of individual instances. The network can focus more on the location and volume of instances than the whole 3D scene, enhancing the 3D instance-aware capability of the proposed depth thickness field and thereby improving detection accuracy.

The main contributions are summarized as follows.
\begin{itemize}[itemsep=3pt,topsep=0pt,parsep=0pt]
\item We investigate the issue of depth-assisted methods as the limited 3D structure-aware capability of existing depth representations and present a network (termed MonoDTF) with a novel depth representation, Depth Thickness Field, to solve it.

\item We design a Scene-Level Depth Retargeting (SDR) module that retargets traditional depth representations to the proposed Depth Thickness Field, incorporating the scene-level perception of 3D structures.

\item To enhance the 3D instance-aware capability of the depth thickness field, an Instance-Level Spatial Refinement (ISR) module is proposed to refine the voxel space and further improve detection accuracy.

\item Compared with SoTA solutions, the proposed MonoDTF demonstrates superior accuracy and generalizability.

\end{itemize}

\begin{figure*}
  \setlength{\belowcaptionskip}{-0.3cm}
  \centering
  \begin{subfigure}{1.0\linewidth}
  \includegraphics[width=1.0\linewidth]{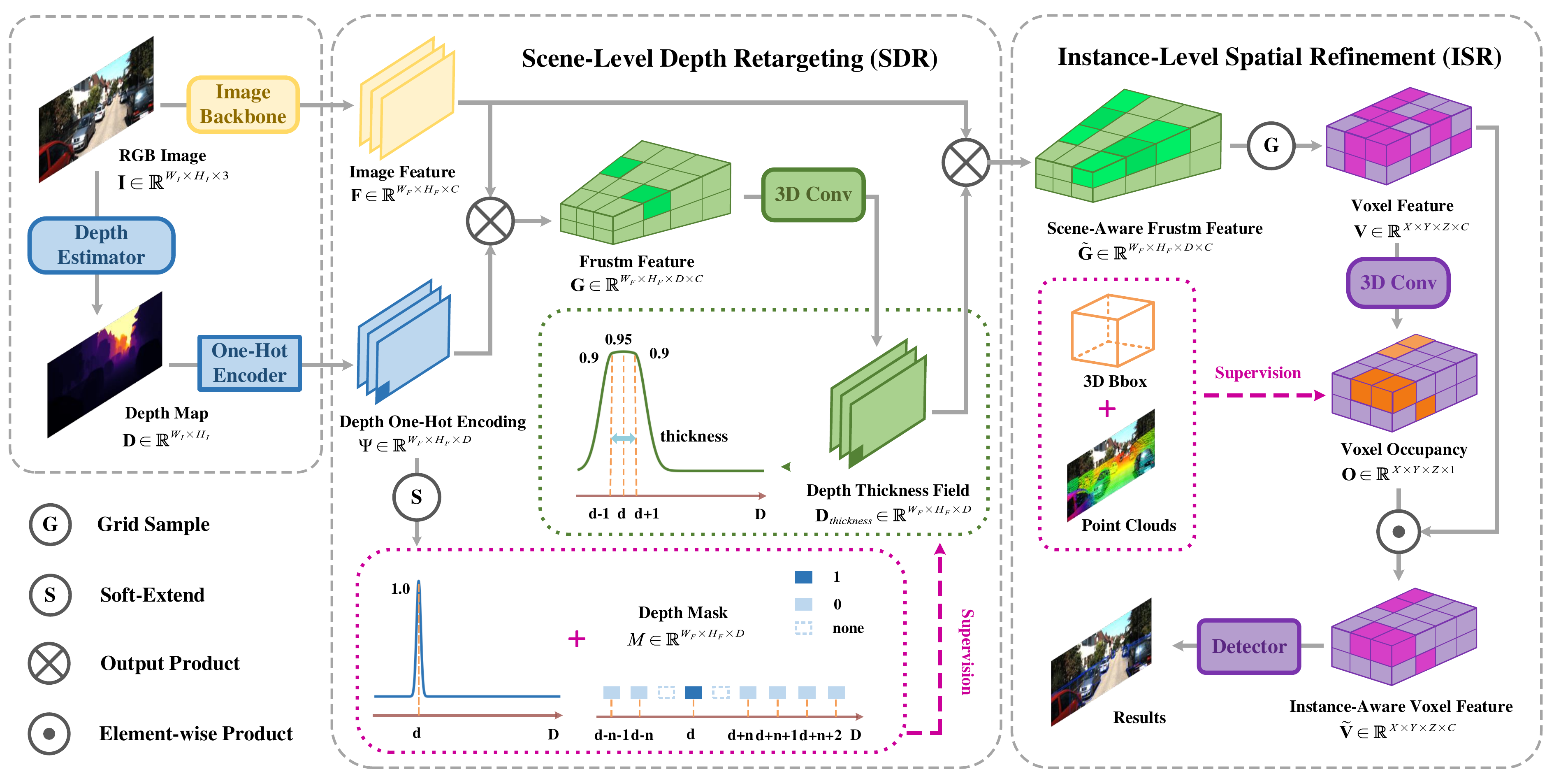}
  \end{subfigure}
  \caption{\textbf{The overall framework of our proposed MonoDTF.} The input image is first sent to the backbone to extract the features and the depth map is transformed to depth one-hot encoding in parallel. The Scene-Level Depth Retargeting (SDR) module implicitly learns 3D scene structure features, retargeting depth one-hot encoding to depth thickness field (Sec. \ref{subsec:SDR}). The Instance-Level Spatial Refinement (ISR) module refines the voxel features with the guidance of instances, eliminating the ambiguity of 3D occupation (Sec. \ref{subsec:ISR}). Finally, the detector is applied to predict the 3D bounding boxes.}
  \label{fig:network}
\end{figure*}

%% file: sec/2_relwork.tex
\section{Related Work}
\label{sec:relwork}

\subsection{Monocular 3D Detection}
In recent years, substantial progress has been made in monocular 3D detection. Nevertheless, inferring 3D information from a 2D image presents an ill-posed problem. Previous works \cite{huang2022monodtr, peng2022did, lu2021geometry, reading2021categorical} have employed a single module for the supervised learning of dense depth maps. DD3D \cite{park2021pseudo} utilizes additional extensive unlabeled data from DDAD15M for depth pre-training, resulting in substantial enhancements in the performance of monocular 3D detection. 
More recently, MonoCD \cite{yan2024monocd} suggests leveraging the geometric relationships among various depth cues to attain complementary form enhancements. Recent progress in monocular 3D detection has incorporated instance attributes \cite{zhou2023monoatt,li2024monolss}, which helps to tackle the inherent difficulties of this ill-posed task more effectively. Our method focuses on retargeting the traditional depth representations to a new formulation: depth thickness field, thereby improving the accuracy of 3D detection.

\subsection{Monocular Depth Estimation}
Monocular depth estimation aims to generate a single depth value for every pixel in an image. Neural network-based methods \cite{wang2024weatherdepth,godard2019digging,shen2022panoformer} often regress continuous depth by aggregating color and geometric structure information from an image. However, these methods are susceptible to overfitting to a single scenario, a common issue in monocular depth estimation. Recent methods \cite{cabon2020virtual,straub2019replica} have effectively mitigated the problem by creating large-scale relative depth datasets to learn relative relations. 
DAMv2 \cite{yang2024depth} unleashes the power of large-scale unlabeled data through semantic segmentation-assisted supervision and data augmentation. Metric3Dv2 \cite{yin2023metric3d} introduces a canonical camera transformation to resolve metric depth ambiguity across different camera setups, significantly improving depth accuracy and generalization. 
In our work, we utilize the depth map generated by the depth estimation model as a depth prior for the 3D detection model, guiding the generation of the depth thickness field representation.
\subsection{3D Spatial Representations}
Representing features in 3D space, such as BEV grids, removes scale differences in the targets and makes it easier for the network to learn feature scale consistency. Many works \cite{li2023bevdepth,li2022bevformer,huang2021bevdet,reading2021categorical,liu2023bevfusion} employ BEV representations and achieve great success. LSS \cite{philion2020lift} generates point clouds by learning a pixel-level depth distribution and projecting it to obtain BEV features. 
BEVDepth \cite{li2023bevdepth} supervises the depth distribution using a projection of the depth map from the point cloud, thus improving the 3D detection performance. 
CaDDN \cite{reading2021categorical} first used BEV representations for monocular 3D detection, and our approach follows its basic architecture design. Additionally, occupancy as an explicit representation of 3D space has recently experienced a surge in research. TPVFormer \cite{huang2023tri} employs sparse 3D occupancy labels from LiDAR as the supervision to obtain 3D features.
Unlike the above methods, we concentrate on combining occupancy representation and 3D bounding box (bbox) to explore the mutual support of occupancy and monocular 3D detection tasks.

%% file: sec/3_method.tex
\section{Methodology}
\label{subsec:thickness}
\begin{figure*}[t]
  \centering
  \begin{subfigure}{0.99\linewidth}
    \includegraphics[width=1.0\linewidth]{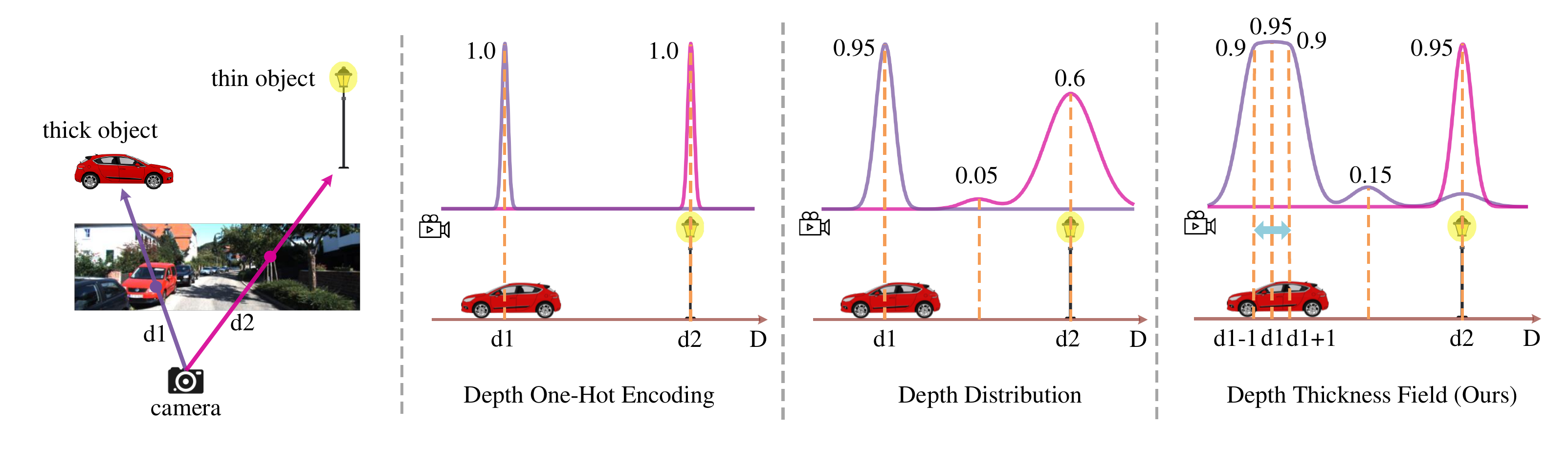}
  \end{subfigure}
  \caption{\textbf{Comparisons between different depth representations.} The two curves indicate the confidence of the pixel on each depth bin, obtained from three different depth representations respectively. The \textcolor[RGB]{115,78,156}{\textbf{purple}} curve represents a thick object, and the \textcolor[RGB]{214,0,147}{\textbf{claret}} one represents a thin object. The depth thickness field has the unique ability to represent the object thickness along the camera ray.}
  \label{fig:resresentation}
\end{figure*}
\subsection{Preliminary and Overview}
\label{subsec:framework}
\textbf{Task Definition.} 
Given a single RGB image, monocular 3D object detection aims to recognize objects of interest and predict their 3D attributes, including 3D location $(x, y, z)$, dimension $(h, w, l)$, and orientation $\theta$.
Existing approaches mainly focus on outdoor dynamic objects, with particular emphasis on the \textit{Car} category \cite{ma20233d}.\\
\textbf{Framework Overview.} As illustrated in Figure \ref{fig:network}, our framework takes a 3-channel RGB image and a single-channel depth map (generated by a pre-trained model) as input and outputs the 3D bounding box parameters of the object in the image. Initially, the RGB image $\mathbf{I}\in \mathbb{R} ^{W_I\times H_I\times 3}$ is passed through the backbone (DLA34 \cite{yu2018deep}) to extract image features, while the depth map $\mathbf{D}\in \mathbb{R} ^{W_I\times H_I}$ is converted into depth one-hot encoding representation. Next, the extracted image features $\mathbf{F}\in \mathbb{R} ^{W_F\times H_F\times C}$ and depth one-hot encoding $\Psi \in \mathbb{R} ^{W_F\times H_F\times D}$ are fed into the SDR module for further processing. Specifically, we employ the depth one-hot encoding representation to transform the extracted features into frustum features format $\mathbf{G}\in \mathbb{R} ^{W_F\times H_F\times D\times C}$. Then the depth thickness field $\mathbf{D}_{thickness}\in \mathbb{R} ^{W_F\times H_F\times D}$ is adaptively learned under the supervision of our proposed soft-extended one-hot encoding. This depth thickness field is then used to refine the original image features and produce new frustum features $\mathbf{\tilde{G}}\in \mathbb{R} ^{W_F\times H_F\times D\times C}$. Following this, grid sampling is applied to voxelize the 3D features. The voxel features, denoted as $\mathbf{V} \in \mathbb{R}^{X \times Y \times Z \times C}$, are subsequently fed into the ISR module. This module refines the voxel space representation by learning an occupancy map, which is guided by the enhanced occupancy labels. 
Finally, the refined voxel features $\mathbf{\tilde{V}}\in \mathbb{R} ^{X\times Y\times Z\times C}$ are compressed into a bird's-eye view (BEV) space and forwarded to the 3D detection head to output the results.

\subsection{Depth Thickness Field}
\label{subsec:dtf}
\begin{figure}[t]
  \centering
  \begin{subfigure}{0.99\linewidth}
    \includegraphics[width=1.0\linewidth]{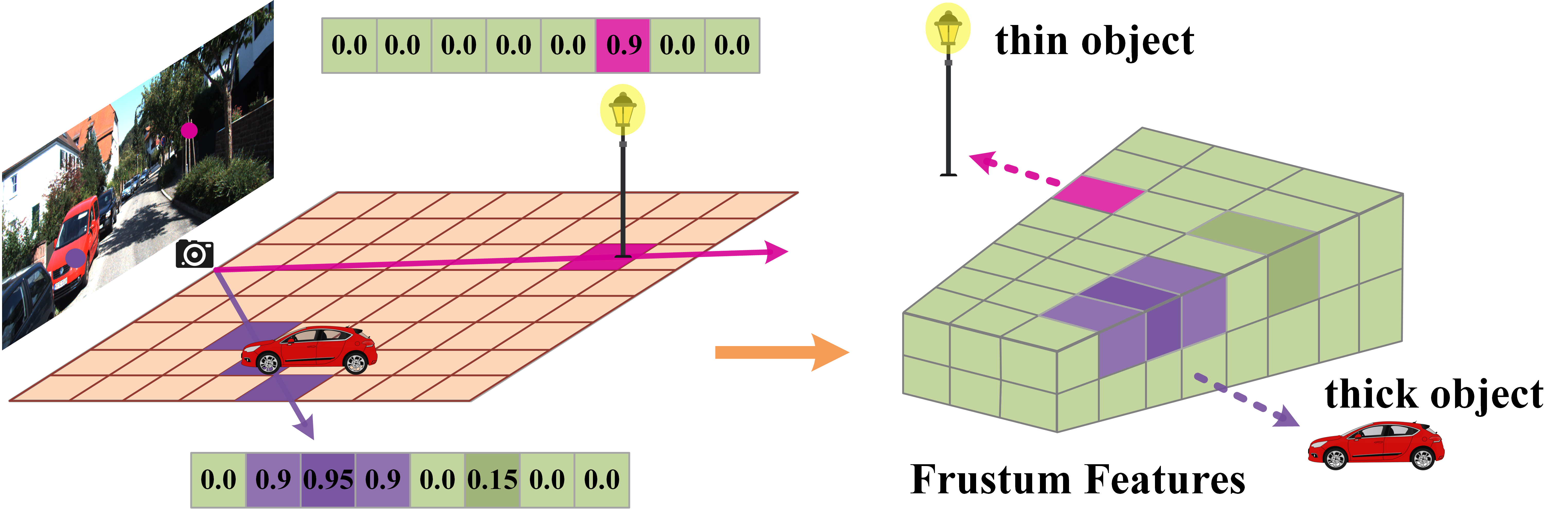}
  \end{subfigure}
  \caption{Thickness indicated by Depth Thickness Field along the camera ray.}
  \label{fig:4}
\end{figure}

Existing 3D object detection methods \cite{peng2022did, yan2024monocd} incorporate depth information to enhance 3D perception. However, depth estimation and 3D object detection are different tasks. Inappropriate depth representations often fail to provide meaningful cues for accurate detection. In this paper, we propose a novel depth representation method, termed \textit{Depth Thickness Field}, which is more effective for 3D object detection. To clarify our contributions, we first review two commonly used depth representation methods—depth one-hot encoding and depth distribution—before detailing our proposed approach.\\
\textbf{Depth One-Hot Encoding.} Given a pixel located at position $(u,v)$ in a depth map, its depth value $z$ can be converted to $z'$ using linear-increasing discretization (LID) \cite{tang2020center3d}. Then we employ $D$ depth bins to represent the corresponding depth one-hot encoding of that pixel, which can be expressed as: 
\begin{align} 
    \Psi(u,v) &= \left[ \phi_1, \phi_2, \cdots,  \phi_d, \cdots,\phi_D \right], 
    \label{eq:1}
\end{align}
where 
\begin{equation}
\phi_d=\left\{ \begin{array}{c}
	1 \quad if\,\,d=z',\\
	0 \quad if\,\,d\ne z'.\\
\end{array} \right. 
  \label{eq:2}
\end{equation}
By applying Eq. \ref{eq:1} to each pixel in the depth map, we can obtain the depth one-hot encoding of the entire map, denoted as $\Psi$. This representation is sensitive to depth variations, aiding in distinguishing different objects. However, it heavily depends on the accuracy of the depth map. Moreover, this method generates a narrow response on the surface of the 3D object, limiting its ability to fully capture the three-dimensional structure.\\
\textbf{Depth Distribution.} Similar to depth one-hot encoding representation, the depth distribution of a pixel in the depth map can be represented as follows:
\begin{equation}
P = \left[ p_1,p_2, \cdots , p_d, \cdots , p_D \right], \\
  \label{eq:3}
\end{equation}
where \(p_d\) represents the probability that the depth value \(z'\) lies within the \(d\)-th depth bin, satisfying the following constraint:
\begin{equation}
\sum_{d=1}^D{p_d}=1 , \quad p_d\geqslant 0.\\
  \label{eq:4}
\end{equation}
Compared to depth one-hot encoding, depth distribution incorporates uncertainty when representing the depth probability distribution with depth bins, reducing the reliance on precise depth estimation. However, it still fails to fully capture the 3D structure (e.g., thickness) due to the absence of implicit thickness-aware guidance.\\
\textbf{Depth Thickness Field.} Unlike the two depth representations discussed above, we propose a novel depth thickness field method for mapping the given depth map. Our approach can implicitly represent the thickness of objects (as shown in Fig. \ref{fig:4}), thereby enabling meaningful depth cues for 3D object detection. The details are as follows.

Given a pixel in a depth map, its depth thickness field \(F\) can be expressed as:
\begin{equation}
F = \left[ f_1, f_2, \cdots , f_d,\cdots, f_D \right], \\
  \label{eq:5}
\end{equation}
where $0\leqslant f_d\leqslant 1$, represents the value of the \(d\)-th depth bin along the camera ray.

The depth thickness field is not directly derived from the depth map. Instead, it is adaptively regressed by the network. While the depth distribution is also regressed, its supervision depends on pixel-wise ground truth depth values, which results in a limited range of responses within the depth bins. In contrast, our depth thickness field regression is guided by a soft-extended depth one-hot encoding constraint (detailed in Sec. \ref{subsec:SDR}), producing a broader range of responses (as shown in Fig. \ref{fig:resresentation}) and enabling an adaptive representation of object thickness.

\subsection{Scene-Level Depth Retargeting}
\label{subsec:SDR}

  
\begin{figure}[t]
\setlength{\belowcaptionskip}{-0.15cm}
  \centering
  \begin{subfigure}{0.49\linewidth}
    \includegraphics[width=1.0\linewidth]{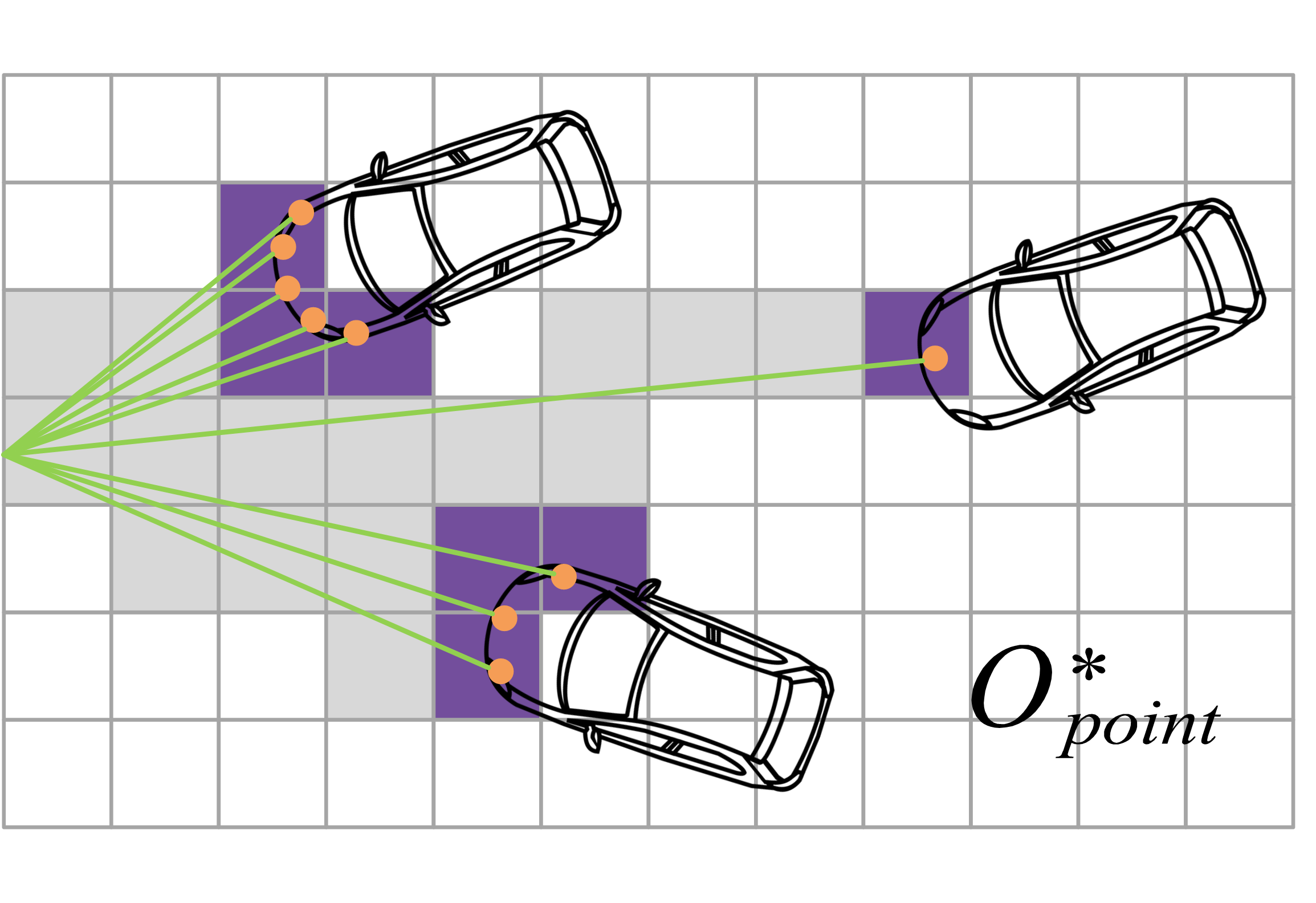}
    \caption{Occupancy labels obtained from OccpancyM3D \cite{peng2024learning}.}
    \label{fig:5a}
  \end{subfigure}
  \begin{subfigure}{0.49\linewidth}
    \includegraphics[width=1.0\linewidth]{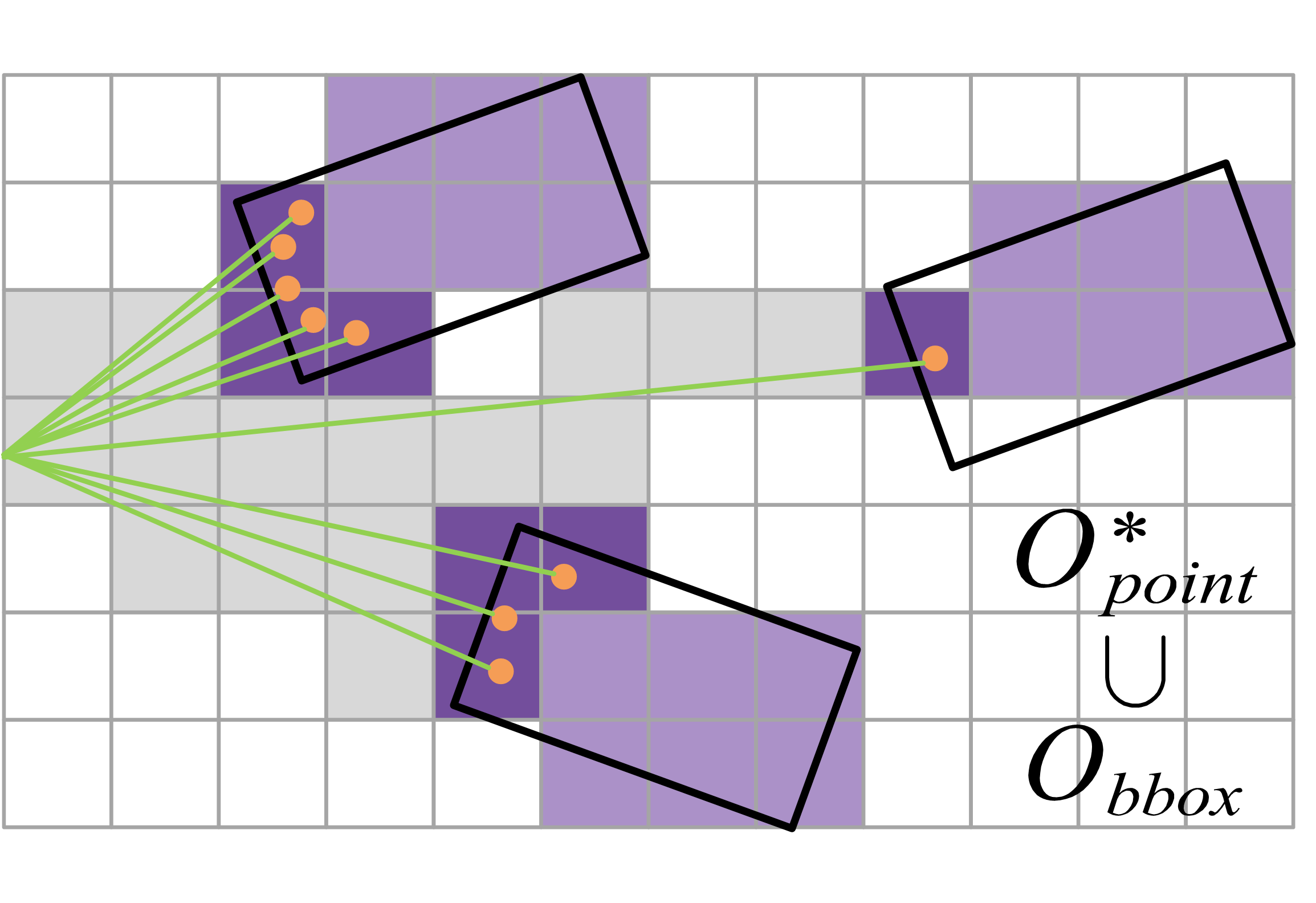}
    \caption{Occupancy labels obtained from MonoDTF (Ours).}
    \label{fig:5b}
  \end{subfigure}
  \caption{  
  An slice example of occupancy map at a certain height. (a) OccupancyM3D takes voxels containing point clouds as positive samples (\textcolor[RGB]{115,78,156}{\textbf{dark purple}}) and those passed through by camera ray and without point clouds as negative samples (\textcolor[RGB]{125,125,125}{\textbf{gray}}). (b) Our method enhances (a) by incorporating voxels contained within the 3D boxes as positive samples (\textcolor[RGB]{171,145,200}{\textbf{light purple}}).
  }
  \label{fig:5}
\end{figure}
These methods \cite{reading2021categorical, peng2024learning} based on the LSS framework usually use depth information to explicitly transform 2D image features into 3D frustum features. However, due to the imperfect depth representation, the generated frustum features are not sufficient to perceive the 3D structure, resulting in inaccurate detection results. To address the challenges, we design the Scene-Level Depth Retargeting (SDR) module to retarget the traditional depth representations to the depth thickness field, thereby obtaining more meaningful frustum features for the subsequent 3D detection procedure.
Specifically, given an RGB image \(\mathbf{I} \in \mathbb{R}^{W_I \times H_I \times 3}\) and its corresponding depth map \(\mathbf{D} \in \mathbb{R}^{W_I \times H_I}\), we first use a pre-trained backbone to extract image features \(\mathbf{F} \in \mathbb{R}^{W_F \times H_F \times C}\). Simultaneously, we downsample the depth map and apply the LID method to produce $D$ depth bins for each pixel in the depth map. Subsequently, we employ Eq. \ref{eq:1} to obtain the depth one-hot encoding representation \(\Psi \in \mathbb{R}^{W_F \times H_F \times D}\) of the depth map.

Given a pixel located at the position $(u,v)$ in the feature map, we can get the corresponding frustum feature grid \(\textbf{G}(u,v)\) as follows:
\begin{equation}
\textbf{G}\left( u,v \right) =\Psi\left( u,v \right) \otimes \textbf{F}\left( u,v \right), \\
  \label{eq:6}
\end{equation}
where \(\textbf{G}\left( u,v \right)\) $\in \mathbb{R}^{D \times C}$; $\otimes$ indicates the outer products. Applying Eq. \ref{eq:6} to each pixel in the feature map, we can get the whole frustum features $\textbf{G} \in \mathbb{R} ^{W_F\times H_F\times D\times C}$. Then we use three 3D convolution layers followed by a sigmoid function to refine the frustum features and regress the depth thickness field \(\textbf{D}_{thickness}\in \mathbb{R} ^{W_F\times H_F\times D}\). Finally, we can produce meaningful and progressive frustum features $\tilde{\textbf{G}}$ as follows:
\begin{equation}
\tilde{\textbf{G}} =\textbf{D}_{thickness} \otimes \textbf{F}, \\
  \label{eq:7}
\end{equation}
where \(\tilde{\textbf{G}}\in \mathbb{R} ^{W_F\times H_F\times D\times C}\).

As previously mentioned, the depth thickness field is adaptively generated by the network. To ensure the DTF accurately reflects object thickness, we impose the following constraints.
Firstly, we define a mask to extend the response range within the standard depth one-hot encoding:
\begin{equation}
    M = [m_1, m_2, \cdots, m_d, \cdots, m_D],
    \label{eq:8}
\end{equation}
where 
\begin{equation}
m_d=\left\{ \begin{array}{rl}

	1 &\quad |d-z'|>l \, or \, d=z',\\
	0 &\quad else.\\
\end{array} \right. 
  \label{eq:9}
\end{equation}
where $l$ represents the extended response range; $z'$ has been defined in Sec. \ref{subsec:dtf}.
Then the soft-extended depth one-hot encoding $\tilde{\Psi}(u,v)$ of a pixel in the depth map can be denoted as follows:
\begin{equation}
    \tilde{\Psi}(u,v) = \Psi(u,v)\odot M,
    \label{eq10}
\end{equation}
where $\odot$ indicates the element-wise products. 
Finally, we introduce the focal loss \cite{lin2017focal} to formulate the objective function of this part as follows:
\begin{equation}
L_{thickness}=\frac{1}{W_F\cdot H_F}FL(\tilde{\Psi}, \textbf{D}_{thickness}).\\
  \label{eq:11}
\end{equation}

\subsection{Instance-Level Spatial Refinement}
\label{subsec:ISR}
\begin{figure}[t]
\setlength{\abovecaptionskip}{-0.1cm}
\setlength{\belowcaptionskip}{-0.15cm}
  \centering
  \begin{subfigure}{0.99\linewidth}
    \includegraphics[width=1.0\linewidth]{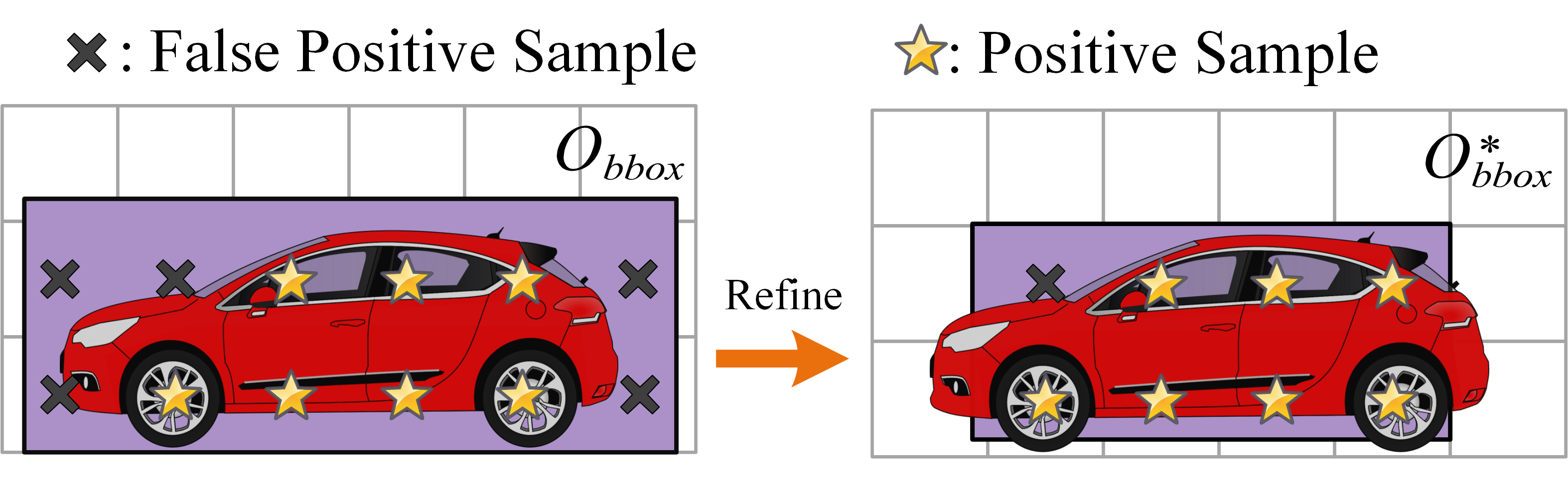}
    \label{figocc3}
  \end{subfigure}
  
  \caption{An slice example of occupancy map. 
  Refining the 3D bounding box can reduce the False Positive Occupancy, thereby alleviating the ambiguity  in the occupancy map.}  
  \label{fig:6}
\end{figure}

\begin{table*}[t]
\setlength{\belowcaptionskip}{-0.3cm}
\centering
\footnotesize
\begin{tabular}{c|c|ccc|ccc|ccc}
\toprule
\multirow{2}{*}{Method} & \multirow{2}{*}{Reference} & \multicolumn{3}{c|}{$Test, AP_{3D}|R_{40}$} & \multicolumn{3}{c|}{$Test, AP_{BEV}|R_{40}$} & \multicolumn{3}{c}{$Val, AP_{3D}|R_{40}$}\\
& & Easy & Moderate& Hard & Easy& Moderate&Hard& Easy& Moderate&Hard\\
\midrule  
MonoDTR \cite{huang2022monodtr} & \textit{CVPR22} & 21.99 & 15.39 & 12.73 & 28.59 & 20.38 & 17.14 & 28.84 & 20.61 & 16.38\\  
DEVIANT \cite{kumar2022deviant} & \textit{ECCV22} & 21.88 & 14.46 & 11.89 & 29.65 & 20.44 & 17.43 & 24.63 & 16.54 & 14.52\\  
DID-M3D \cite{peng2022did} & \textit{ECCV22} & 24.40 & 16.29 & 13.75 & 32.95 & 22.76 & 19.83 & 22.98 & 16.12 & 14.03\\
MonoDDE \cite{li2022diversity} & \textit{CVPR22} & 24.93 & 17.14 & 15.10 & 33.58 & 23.46 & 20.37 & 26.66 & 19.75 & 16.72\\  
MonoNeRD \cite{xu2023mononerd} & \textit{ICCV23} & 22.75 & 17.13 & 15.63 & 31.13 & 23.46 & 20.97 & 22.03 & 15.44 & 13.99\\
MonoDETR \cite{zhang2023monodetr} & \textit{ICCV23} & 25.00 & 16.47 & 13.58 & 33.60 & 22.11 & 18.60 & 28.84 & 20.61 & 16.38\\  
MonoUNI \cite{jinrang2023monouni} & \textit{NeurIPS23} & 24.75 & 16.73 & 13.49 & 33.28 & 23.05 & 19.39 & 24.51 & 17.18 & 14.01\\
MonoATT \cite{zhou2023monoatt} & \textit{CVPR23} & 24.72 & 17.37 & 15.00 & \textcolor{blue}{\textbf{36.87}} & 24.42 & 21.88 & 29.01 & \textcolor{blue}{\textbf{23.49}} & 19.60\\
FD3D \cite{wu2024fd3d}& \textit{AAAI24} & 26.35 & 18.72 & 15.97 & 34.20 & 23.72 & 20.76 & 28.22 & 20.23 & 17.04 \\
MonoLSS \cite{li2024monolss} & \textit{3DV24} & 26.11 & 19.15 & 16.94 & 34.89 & \textcolor{blue}{\textbf{25.95}} & \textcolor{blue}{\textbf{22.59}} & 25.91 & 18.29 & 15.94\\
MonoCD \cite{yan2024monocd} & \textit{CVPR24} & 25.53 & 16.59 & 14.53 & 33.41 & 22.81 & 19.57 & 26.45 & 19.37 & 16.38\\  
MonoDiff \cite{ranasinghe2024monodiff} & \textit{CVPR24} & \textcolor{blue}{\textbf{30.18}} & \textcolor{blue}{\textbf{21.02}} & \textcolor{blue}{\textbf{18.16}} & - & - & - & \textcolor{blue}{\textbf{32.18}} & 22.02 & \textcolor{blue}{\textbf{19.84}}\\
OccupancyM3D \cite{peng2024learning} & \textit{CVPR24} & 25.55 & 17.02 & 14.79 & 35.38 & 24.18 & 21.37 & 26.87 & 19.96 & 17.15\\
MonoMAE \cite{jiang2024monomae} & \textit{NeurIPS24} & 25.60 & 18.84 & 16.78 & 34.15 & 24.93 & 21.76 & 30.29 & 20.90 & 17.61\\ 
\midrule
\textbf{MonoDTF(Ours)} & - &  \textcolor{red}{\textbf{32.02}} & \textcolor{red}{\textbf{21.19}} & \textcolor{red}{\textbf{18.80}} &\textcolor{red}{\textbf{42.67}} & \textcolor{red}{\textbf{28.92}} & \textcolor{red}{\textbf{25.89}} & \textcolor{red}{\textbf{36.17}} & \textcolor{red}{\textbf{26.76}} & \textcolor{red}{\textbf{23.67}}\\
\textit{Improvement} & \textit{v.s. second-best} & \textcolor{blue}{+1.84} & \textcolor{blue}{+0.17} & \textcolor{blue}{+0.64} & \textcolor{blue}{+5.80} & \textcolor{blue}{+2.97} & \textcolor{blue}{+3.3} & \textcolor{blue}{+3.99} & \textcolor{blue}{+3.27} & \textcolor{blue}{+3.83}\\   
\bottomrule
\end{tabular}
\caption{Comparisons on KITTI \textit{test} and \textit{val} sets for Car category. Results are shown using the $AP(IoU=0.7)|R_{40}$ metric. We indicate the highest result with \textcolor{red}{\textbf{red}} and the second highest with \textcolor{blue}{\textbf{blue}}. The performance metrics for the other methods are reported from the respective published results.}
\label{tab1}
\end{table*}
OccupancyM3D \cite{peng2024learning} enhances 3D spatial features by learning an occupancy representation in voxel space. However, the labels used to guide the occupancy representation generation emanate from sparse point clouds, which are confined to the object's surface (as shown in Fig. \ref{fig:5a}). As a result, they fail to accurately capture the thickness along the camera ray. Additionally, the limited perception range of LiDAR further restricts the ability to impose effective constraints on distant objects, exacerbating the ill-posed problem in the labels. This limitation significantly impacts the network’s performance. To address this challenge, we design the Instance-Level Spatial Refinement (ISR) module to refine the voxel spatial representation with the guidance of instances.

Given voxel features $\textbf{V}\in \mathbb{R} ^{X\times Y\times Z\times C}$, we use 3D convolution with a sigmoid function to obtain the voxel occupancy features $\textbf{O}\in \mathbb{R} ^{X\times Y\times Z\times 1}$ which then are employed to refine the previous voxel features. 
This procedure can be expressed as follows:
\begin{equation}
\begin{aligned}
  \textbf{O} &= \text{Sigmoid}\left( f\left( \textbf{V} \right) \right), \\
  \tilde{\textbf{V}} &= \textbf{O} \odot \textbf{V},
\end{aligned}
\label{eq:12}
\end{equation}
\noindent where $f(\cdot)$ denotes a 3D hourglass-like design \cite{chang2018pyramid} and $\odot$ denotes the element-wise products.\\ 
\textbf{Occupancy Labels}. 
To generate more complete and precise occupancy labels for 3D object detection, we propose leveraging 3D bounding boxes to assist sparse point clouds in representing object thickness. 
Firstly, similar to OccupancyM3D \cite{peng2024learning}, we denote the 3D occupancy labels derived from sparse point clouds as \( O_{point}^* \in \mathbb{R}^{X \times Y \times Z} \), where \( X \), \( Y \), and \( Z \) are determined by the predefined voxel size and detection range, as illustrated in Figure \ref{fig:5a}. Then we denote an additional 3D occupancy label derived from 3D bounding boxes as \( O_{bbox} \in \mathbb{R}^{X \times Y \times Z} \) to complement \( O_{point}^*\), as shown in Figure \ref{fig:5b}.

Considering that the ISR module learns the occupancy representation within the voxel space and that the original 3D bounding boxes do not precisely conform to the object’s shape, directly using them would result in ambiguous occupancy labels. To address this issue, we propose constructing refined 3D bounding boxes, derived from the originals, as shown in Fig. \ref{fig:6}. These adjusted boxes retain the same center but are scaled down to reduce ambiguity in the occupancy labels.
Specifically, let \( \tilde{\textbf{B}} \in \mathbb{R}^{N \times 7} \) represent the refined 3D bounding boxes. For each voxel block within the voxel space, if it falls within any of the 3D bounding boxes, we designate it as occupied. Using this approach, we obtain additional 3D occupancy labels \( O_{bbox}^* \in \mathbb{R}^{X \times Y \times Z} \) derived from the 3D bounding boxes. The final occupancy label is then obtained by taking the union of \( O_{point}^* \) and \( O_{bbox}^* \), i.e., \( O^* = O_{point}^* \cup O_{bbox}^* \).

The ISR module enhances the perception of instances by adding occupancy labels from the 3D bounding boxes and then using it to refine the voxel spatial representation, which is beneficial in improving detection accuracy.

\subsection{Loss Function}
Our objective function consists of three components: the first for 3D object detection, the second for learning occupancy in 3D voxel space, and the third for learning the depth thickness field. The first component follows the approach of CaDDN \cite{reading2021categorical}, denoted as $\mathcal{L}_{org}$. For the occupancy component, we adopt the loss function from OccupancyM3D \cite{peng2024learning}, represented as $\mathcal{L}_{occ}$. The third component involves the depth thickness field, for which we use the thickness loss $\mathcal{L}_{thickness}$, as defined in Section \ref{subsec:SDR}. The overall objective function is then formulated as follows:
\begin{equation}
\mathcal{L} =\mathcal{L} _{org}+ \mathcal{L} _{occ}+ \mathcal{L} _{thickness}\\
\label{eq:13}
\end{equation}

%% file: sec/4_exps.tex
\section{Experiments}
\begin{table*}[t]
\centering
\footnotesize
\begin{tabular}{c|c|cccc|cccc}
\toprule
\multirow{2}{*}{Method} & \multirow{2}{*}{Reference} & \multicolumn{4}{c|}{$3D mAP/mAPH(IoU=0.7)$} & \multicolumn{4}{c}{$3D mAP/mAPH(IoU=0.5)$} \\
& & Overall & 0 - 30m & 30 - 50m & 50m - $\infty$& Overall & 0 - 30m & 30 - 50m & 50m - $\infty$\\
\midrule  
\multicolumn{10}{c}{\multirow{1}{*}{LEVEL 1}}\\
\midrule  
DEVIANT \cite{kumar2022deviant} & \textit{ECCV22} & 2.69/2.67 & 6.95/6.90 & 0.99/0.98 & 0.02/0.02 & 10.98/10.89 & 26.85/26.64 & 5.13/5.08 & 0.18/0.18\\
DID-M3D \cite{peng2022did} & \textit{ECCV22} & -/- & -/- & -/- & -/- & 20.66/20.47 & 40.92/40.60 & 15.63/15.48 & 5.35/5.24\\
MonoNeRD \cite{xu2023mononerd} & \textit{ICCV23} & \textcolor{blue}{\textbf{10.66/10.56}} & 27.84/27.57 & \textcolor{blue}{\textbf{5.40/5.36}} & \textcolor{red}{\textbf{0.72/0.71}} & 31.18/30.70 & 61.11/60.28 & \textcolor{blue}{\textbf{26.08/25.71}} & \textcolor{blue}{\textbf{6.60/6.47}}\\
MonoUNI \cite{jinrang2023monouni} & \textit{NeurIPS23} & 3.20/3.16 & 8.61/8.50 & 0.87/0.86 & 0.13/0.12 & 10.98/10.73 & 26.63/26.30 & 4.04/3.98 & 0.57/0.55\\
MonoLSS \cite{li2024monolss} & 3DV24 & 3.71/3.69 & 9.82/9.75 & 1.14/1.13 & 0.16/0.16 & 13.49/13.38 & 33.64/33.39 & 6.45/6.40 & 1.29/1.26\\
MonoDiff \cite{ranasinghe2024monodiff} & \textit{CVPR24} & -/- & -/- & -/- & -/- & \textcolor{blue}{\textbf{32.28/31.4}9} & \textcolor{blue}{\textbf{63.94/62.13}} & 25.91/25.47 & \textcolor{red}{\textbf{7.51/7.34}}\\ 
OccupancyM3D \cite{peng2024learning} & \textit{CVPR24} & 10.61/10.53 & \textcolor{blue}{\textbf{29.18/28.96}} & 4.49/4.46 & 0.41/0.40 & 28.99/28.66 & 61.24/60.63 & 23.25/23.00 & 3.65/3.59\\
\midrule
MonoDTF(Ours) & - & \textcolor{red}{\textbf{18.04/17.79}} & \textcolor{red}{\textbf{47.57/46.96}} & \textcolor{red}{\textbf{9.09/9.00}} & \textcolor{blue}{\textbf{0.64/0.63}} & \textcolor{red}{\textbf{42.06/41.24}} & \textcolor{red}{\textbf{80.81/79.46}} & \textcolor{red}{\textbf{40.42/39.81}} & 6.41/6.21\\
\midrule
\multicolumn{10}{c}{\multirow{1}{*}{LEVEL 2}}\\
\midrule  
DEVIANT \cite{kumar2022deviant} & \textit{ECCV22} & 2.52/2.50 & 6.93/6.87 & 0.95/0.94 & 0.02/0.02 & 10.29/10.20 & 26.75/26.54 & 4.95/4.90 & 0.16/0.16\\
DID-M3D \cite{peng2022did} & \textit{ECCV22} & -/- & -/- & -/- & -/- & 19.37/19.19 & 40.77/40.46 & 15.18/15.04 & 4.69/4.59\\
MonoNeRD \cite{xu2023mononerd} & \textit{ICCV23} & \textcolor{blue}{\textbf{10.03}}/9.93 & 27.75/27.48 & \textcolor{blue}{\textbf{5.25/5.21}} & \textcolor{red}{\textbf{0.60/0.59}} & 29.29/28.84 & 60.91/60.08 & \textcolor{blue}{\textbf{25.36/25.00}} & \textcolor{blue}{\textbf{5.77/5.66}}\\
MonoUNI \cite{jinrang2023monouni}& \textit{NeurIPS23} & 3.04/3.00 & 8.59/8.48 & 0.85/0.84 & 0.12/0.12 & 10.38/10.24 & 26.57/26.24 & 3.95/3.89 & 0.53/0.51\\
MonoLSS \cite{li2024monolss} & 3DV24 & 3.27/3.25 & 9.79/9.73 & 1.11/1.10 & 0.15/0.15 & 13.12/13.02 & 33.56/33.32 & 6.28/6.22 & 1.15/1.13\\
MonoDiff \cite{ranasinghe2024monodiff} & \textit{CVPR24} & -/- & -/- & -/- & -/- & \textcolor{blue}{\textbf{30.73/30.48}} & \textcolor{blue}{\textbf{63.86/62.92}} & 25.28/24.86 & \textcolor{red}{\textbf{6.43/6.29}}\\
OccupancyM3D \cite{peng2024learning} & \textit{CVPR24} & 10.02/\textcolor{blue}{\textbf{9.94}} & \textcolor{blue}{\textbf{28.38/28.17}} & 4.38/4.34 & 0.36/0.36 & 27.21/26.90 & 61.09/60.49 & 22.59/22.34 & 3.18/3.13\\
\midrule
MonoDTF(Ours) & - & \textcolor{red}{\textbf{16.77/16.54}} & \textcolor{red}{\textbf{46.94/46.34}} & \textcolor{red}{\textbf{8.76/8.67}} & \textcolor{blue}{\textbf{0.55/0.54}} & \textcolor{red}{\textbf{39.20/38.44}} & \textcolor{red}{\textbf{79.93/78.60}} & \textcolor{red}{\textbf{39.04/38.45}} & 5.56/5.39\\
\bottomrule
\end{tabular}
\caption{Comparisons on Waymo \textit{val} set for \textit{Vehicle} category. We indicate the highest result with \textcolor{red}{\textbf{red}} and the second highest with \textcolor{blue}{\textbf{blue}}. The performance metrics for the other methods are reported from the respective published results.}
\label{tab2}
\end{table*}

\begin{table*}[t]
\setlength{\belowcaptionskip}{-0.3cm}
\centering
\footnotesize
\begin{tabular}{c|cc|ccc|ccc}
\toprule
\multirow{2}{*}{Exp.} & \multirow{2}{*}{SDR} & \multirow{2}{*}{ISR} & \multicolumn{3}{c|}{$Test, AP_{3D}|R_{40}/AP_{BEV}|R_{40}$} & \multicolumn{3}{c}{$Val, AP_{3D}|R_{40}/AP_{BEV}|R_{40}$} \\
& & & Easy & Moderate& Hard & Easy & Moderate& Hard\\
\midrule
(a) &  &  & 26.95/37.64 & 16.15/23.26 & 13.45/19.77 & 28.15/41.71 & 21.62/31.53 & 19.11/28.34\\

(b) & \checkmark &  & 31.12/40.76 & 20.04/27.49 & 17.33/24.29 & \textbf{36.71/51.60} & 25.52/35.12 & 22.17/31.83 \\
(c) &  & \checkmark & 31.97/42.29 & 19.79/27.28 & 17.38/24.27 & 35.67/48.06 & 24.96/33.60 &21.50/29.49\\
(d) & \checkmark & \checkmark & \textbf{32.02/42.67} & \textbf{21.19/28.92} & \textbf{18.80/25.92} & 36.17/47.42 & \textbf{26.76/35.68} & \textbf{23.67/31.94}\\
\bottomrule
\end{tabular}
\caption{Main ablations on KITTI \textit{test} and \textit{val} sets. ``Exp." means experiment; ``SDR" refers to Scene-Level Depth Retargeting module; ``ISR" denotes Instance-Level Spatial Refinement module.}
\label{tab3}
\end{table*}


\subsection{Implementation Details}
We employ PyTorch \cite{paszke2019pytorch} for implementation. The network is trained on 3 NVIDIA 4090 (24G) GPUs. We use Adam \cite{kingma2014adam} optimizer with an initial learning rate of 0.001 and employ the one-cycle learning rate policy \cite{smith2018disciplined}. On KITTI dataset \cite{geiger2012we}, we train the model for 80 epochs with a total batch size of 12 and use [2, 46.8] $\times$ [-30.08, 30.08] $\times$ [-3, 1] (meter) for the range of BEV space. On Waymo dataset \cite{ettinger2021large}, we downsample the input size to 640 \(\times\) 960 and train 12 epochs with a detection range [2, 59.6] \(\times\) [-25.6, 25.6] \(\times\) [-2, 2] (meter). The voxel size for the KITTI and Waymo datasets is [0.16, 0.16, 0.16] (meter).
\subsection{Benchmarks and Metrics}
\textbf{KITTI. } KITTI is the most widely used benchmark \cite{geiger2012we} for 3D object detection, consisting of 7,481 training samples and 7,518 testing samples. Only annotations of the training set are made public. The training set is commonly divided into KITTI \textit{train} with 3712 samples and KITTI \textit{val} with 3769 samples following \cite{chen20173d}.
There are three difficulty levels, easy, moderate, and hard, defined by bounding box height and occlusion level. Following common practice, we use $AP_{3D}|_{R_{40}}$ and $AP_{BEV}|_{R_{40}}$ under \textit{IoU} threshold of 0.7 to evaluate the performance.\\
\textbf{Waymo.} The Waymo Open Dataset \cite{ettinger2021large} is a large dataset that contains 798 sequences for training and 202 sequences for validation. Following previous works \cite{reading2021categorical, peng2024learning}, we report the performance on Waymo \textit{val} set for the vehicle category using samples only from the front camera. Results on Waymo val are measured by official evaluation of the mean average precision (mAP) and the mean average precision weighted by heading (mAPH).\\
\begin{table*}[t]
\centering
\footnotesize
\begin{tabular}{c|l|c|ccc|ccc}
\toprule
\multirow{2}{*}{Exp.} & \multirow{2}{*}{Depth Model}& \multirow{2}{*}{SDR\&ISR} & \multicolumn{3}{c|}{$Test, AP_{3D}|R_{40}/AP_{BEV}|R_{40}$} & \multicolumn{3}{c}{$Val, AP_{3D}|R_{40}/AP_{BEV}|R_{40}$}\\
& & & Easy & Moderate& Hard & Easy & Moderate& Hard\\
\midrule
(1) & MonoDepth2* \cite{godard2019digging}& & 20.41/28.33 & 12.03/17.45 & 10.30/14.89 &  20.39/31.34 & 14.90/22.56 & 12.42/19.05\\
(2) & MonoDepth2* & \checkmark & \textbf{21.65/29.71} & \textbf{14.13/19.42} & \textbf{11.73/16.99} & \textbf{22.01/31.57} & \textbf{16.54/23.62} & \textbf{14.17/20.46}\\
\midrule
\multicolumn{3}{c|}{Improvement} & +1.24/+1.38 &  +2.1/+1.97 & +1.43/+2.1 & +1.62/+0.23 & +1.64/+1.06 & +1.75/+1.41\\
\midrule
(3) & DAMv2$\dagger$ \cite{yang2024depth} &  & 26.90/38.51 & 15.40/22.86 & 12.48/19.26 & 35.24/48.36 & 20.97/29.94 & 17.09/24.65\\
(4) & DAMv2$\dagger$ & \checkmark & \textbf{32.74/41.72} & \textbf{20.19/27.11} & \textbf{17.04/23.36} & \textbf{39.82/52.30} & \textbf{27.46/37.79} & \textbf{24.12/33.74}\\
\midrule
\multicolumn{3}{c|}{Improvement}  & +5.84/+3.21 & +4.79/+4.25 & +4.56/+4.10 & +4.58/+3.94 & +6.49/+7.85 & +7.03/+9.09\\
\midrule
(5) & Metric3Dv2 \cite{yin2023metric3d}&  & 26.95/37.64 & 16.15/23.26 & 13.45/19.77 & 28.15/41.71 & 21.62/31.53 & 19.11/28.34\\    
(6) & Metric3Dv2 & \checkmark & \textbf{32.02/42.67} & \textbf{21.19/28.92} & \textbf{18.80/25.89} & \textbf{36.17/47.42} & \textbf{26.76/35.68} & \textbf{23.67/31.94}\\
\midrule
\multicolumn{3}{c|}{Improvement} & +5.25/+5.03 & +5.04/+5.66 & +5.35/+6.12 & +8.02/+5.71 & +5.14/+4.15 & +4.56/+3.60\\
\bottomrule
\end{tabular}
\caption{Ablations of different depth estimation models on KITTI \textit{test} and \textit{val} sets. * denotes the lightweight model and $\dagger$ indicates the further fine-tuned model. ``DAMv2" refers to Depth Anything V2 \cite{yang2024depth}.}
\label{tab5}

\end{table*}

\begin{table}[t]
\centering
\footnotesize
\begin{tabular}{c|c|ccc}
\toprule
\multirow{2}{*}{Exp.} & \multirow{2}{*}{Representation} & \multicolumn{3}{c}{$Test, AP_{3D}|R_{40}/AP_{BEV}|R_{40}$}\\
& & Easy & Moderate& Hard \\
\midrule
1) &  One-Hot & 26.95/37.64 & 16.15/23.26 & 13.45/19.77\\
2) &  Distribution & 30.37/40.42 & 19.16/26.69 & 17.02/24.12\\
3) &  Thickness & \textbf{32.02/42.67} & \textbf{21.19/28.92} & \textbf{18.80/25.89} \\ 
\bottomrule
\end{tabular}
\caption{Ablation of depth representation. One-Hot: Depth One-Hot Encoding; Distribution: Depth Distribution; Thickness: Depth Thickness Field.}
\label{tab6}

\end{table}
\subsection{Main Results}
\textbf{KITTI.} Table \ref{tab1} shows the main results of MonoDTF on KITTI test and val sets. Our method achieves the best results under three-level difficulty settings in $AP_{BEV}$ and $AP_{3D}$. On the test set, MonoDTF exceeds all existing methods and surpasses the second-best under three-level difficulties by +1.84/+0.17/+0.64 in $AP_{3D}$, and by +7.29/+2.97/+3.33 in $AP_{BEV}$. In particular, MonoDTF does not require the depth ground truth (GT) that is hard to obtain for supervision when training the detection model. Guided by accurate depth priors, our method transforms traditional depth representations into a more effective depth thickness field for 3D detection, resulting in a significant improvement over existing approaches.\\
\textbf{Waymo.} We also evaluate our method on Waymo Open Dataset (Waymo) \cite{ettinger2021large}. As shown in Table \ref{tab2}, our method obtains promising results, significantly outperforming existing methods in ``0-30m'', ``30-50m'', and ``Overall''. For example, under LEVEL 1 setting, MonoDTF outperforms the second best by +7.38/+7.23
and +9.78/+9.75 (mAP/mAPH) 
with IoU 0.7 and 0.5 criterions respectively; under LEVEL 2 setting, MonoDTF outperforms the second best by +6.74/+6.60
and +8.47/+7.96 (mAP/mAPH) 
with IoU 0.7 and 0.5 criterions respectively.
This success can be attributed to the effective depth thickness field representation. For objects within [50m, $\infty$], our method performs relatively worse. This is because our method is BEV-based, which has a detection range limitation: [2, 59.6] (\textit{meters}). 

\subsection{Ablations}
\label{ablation}
To illustrate the effectiveness of SDR and ISR modules proposed in this paper, we carry out ablations on both KITTI test and val sets as shown in Table \ref{tab3}. 
\textbf{SDR:} Exp. (a) is the baseline which directly uses depth one-hot encoding to generate 3D features for 3D object detection. In Exp. (b), when retargeting depth representation from depth one-hot encoding to depth thickness field, the detection $AP_{3D}$ increases from 26.95/16.15/13.45 to 31.12/20.04/17.33 on the test set. 
\textbf{ISR:} As shown in Exp. (c), when enforcing the refinement of voxel spatial representation via learning 3D occupation with the guidance of instances, the detection $AP_{3D}$ is boosted to 31.97/19.79/17.38 on the test set. Finally, we can see that MonoDTF obtains +5.25/+5.04/+5.35 in $AP_{3D}$ compared to the baseline of Exp. (d). \\
\textbf{Universality Analysis.} 
In Table \ref{tab5}, we report the performance of MonoDTF when equipped with different depth estimation models. 
DAMv2 and Metric3Dv2 are large models used for depth estimation, distinguished by their methodology. DAMv2 initially produces relative depth, which requires fine-tuning of the metric depth data for 3D detection. In contrast, Metric3Dv2 provides metric depth directly, obviating the requirement for fine-tuning.
The results indicate significant improvements in detection performance when integrated with our method. 
As an early lightweight model for depth estimation, Monodepth2 still benefits from our method for monocular 3D object detection. These experiments demonstrate both the effectiveness and the universality of MonoDTF.\\
\textbf{Ablations on Depth Representations.} In Table \ref{tab6}, we report the performances when replacing the depth thickness field with traditional depth representations (\textit{e.g.}, depth one-hot encoding and depth distribution). We can observe that the depth thickness field performs better than the traditional depth representations. 

%% file: sec/5_conclusion.tex
\section{Conclusion}
\label{sec:conclusion}
In this paper, we propose a depth-adapted network for monocular 3D object detection. Unlike current approaches, we reformulate a new key factor of the depth-assisted method from depth accuracy to the adaptability of depth representation and introduce a novel depth representation. The proposed SDR module retargets traditional depth representations to depth thickness field with self-supervision. We also introduce the ISR module to refine the voxel spatial representation with the guidance of instances. We conduct experiments on the challenging KITTI and Waymo open datasets. The results demonstrate our superiority to existing state-of-the-art (SoTA) methods and the universality when equipped with different depth estimation models. 